\pdfoutput=1
\documentclass[11pt]{article}

\usepackage[]{acl}

\usepackage{times}
\usepackage{latexsym}

\usepackage[T1]{fontenc}
\usepackage[utf8]{inputenc}

\usepackage{microtype}

\usepackage{graphicx}
\usepackage{subfigure}
\usepackage{multirow}
\usepackage{amsmath}
\usepackage{amssymb}
\usepackage{booktabs}

\title{A Multi-turn Machine Reading Comprehension Framework with Rethink Mechanism for Emotion-Cause Pair Extraction}

  \author{Changzhi Zhou, \  Dandan Song\thanks{~ Corresponding Author}, \  Jing Xu, \ Zhijing Wu\\
  School of Computer Science and Technology,\\ Southeast Academy of Information Technology, \\Beijing Institute of Technology, Beijing, China \\% $^{2}$ Beijing Institute of Technology, Fujian, China \\ 
  \texttt{zhou\_changzhi97@163.com} \\ \texttt{sdd@bit.edu.cn}, \  \texttt{xujing@bit.edu.cn} \\ \texttt{wuzhijing.joyce@gmail.com}}

\begin{document}
\maketitle

\begin{abstract}
Emotion-cause pair extraction (ECPE) is an emerging task in emotion cause analysis, which extracts potential emotion-cause pairs from an emotional document. Most recent studies use end-to-end methods to tackle the ECPE task. However, these methods either suffer from a label sparsity problem or fail to model complicated relations between emotions and causes. Furthermore, they all do not consider explicit semantic information of clauses. To this end, we transform the ECPE task into a document-level machine reading comprehension (MRC) task and propose a Multi-turn MRC framework with Rethink mechanism (MM-R). Our framework can model complicated relations between emotions and causes while avoiding generating the pairing matrix (the leading cause of the label sparsity problem). Besides, the multi-turn structure can fuse explicit semantic information flow between emotions and causes.  %In every turn, the interaction between queries and clauses makes full use of explicit semantic information of clauses to improve the performance of emotion or cause extraction. Specifically, a set of candidate emotion-cause pairs is obtained in the first two turns and the rethink mechanism in the third turn circularly verifies each candidate emotion-cause pair. 
Extensive experiments on the benchmark emotion cause corpus demonstrate the effectiveness of our proposed framework, which outperforms existing state-of-the-art methods.\footnote{Data and code are available at https://github.com/zhoucz97/ECPE-MM-R}
\end{abstract}

\section{Introduction}

\begin{figure}[htb]
		\centering
		\subfigure[An example of the ECPE task]
		{
			\label{fig1:sub1}
			\includegraphics[width=1.0\columnwidth]{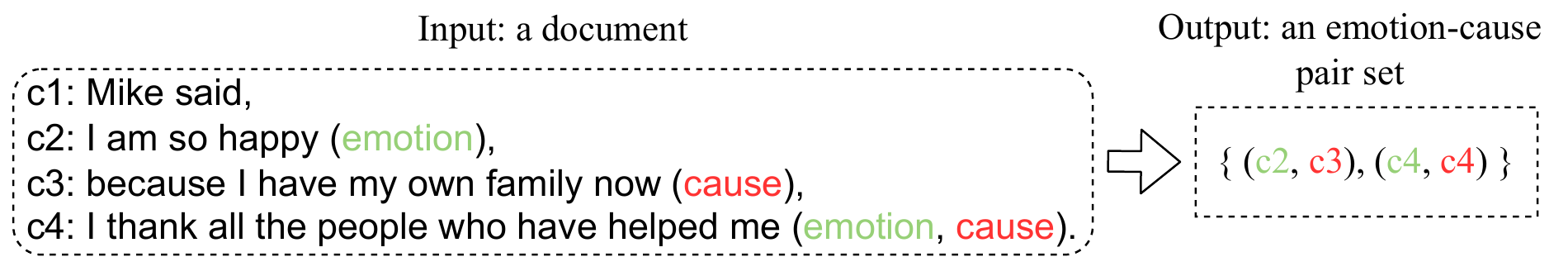}
		}
		\subfigure[Pairing matrix]
		{
			\label{fig1:sub2}
			\includegraphics[width=0.5\linewidth]{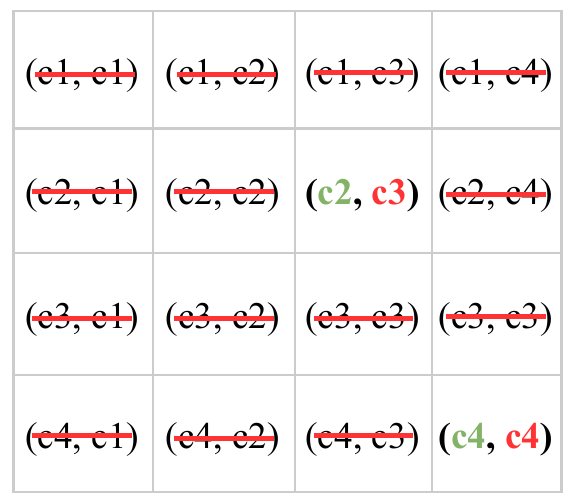}
		}
		\subfigure[Our approach]
		{
			\label{fig1:sub3}
			\includegraphics[width=0.3\linewidth]{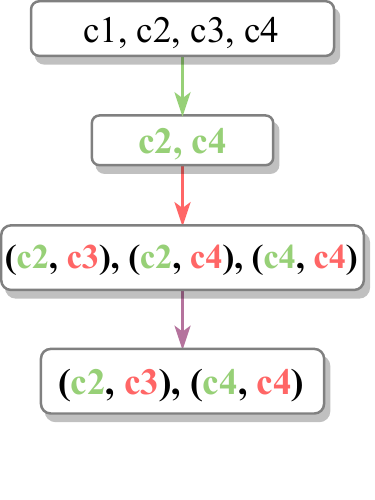}
		}
		\caption{The green colour denotes an emotion clause, and the red colour denotes a cause clause. Figure (a) is an example of the ECPE task. %Given a document, the output of the ECPE task is a set of emotion-cause pairs. 
		Figure (b) is a pairing matrix generated by pair-level end-to-end approaches. Only (c2, c3) and (c4, c4) are valid pairs. Figure (c) shows the processing results by our MM-R in each turn. %The first two turns yield the candidate emotion-cause pair set \{(c2, c3), (c2, c4), (c4, c4)\}. In the third turn, the invalid pair (c2, c4) is abandoned by the rethink mechanism. 
		}
		\label{fig:intro}
	\end{figure}
	
	Emotion cause extraction (ECE) is a classical emotion cause analysis task that aims to extract the corresponding causes of the given emotional expressions in an emotional document ~\cite{gui2016event}. However, the ECE task is not practical in real-world scenarios without annotated emotions. To overcome the limitation and capture mutual indications of emotions and causes together, \citet{xia2019emotion} came up with a new task called emotion-cause pair extraction (ECPE), which aims to extract a potential emotion-cause pair set consisting of all emotions and their corresponding causes from a document. As shown in Figure \ref{fig1:sub1}, c2 is an emotion clause, c3 is the corresponding cause clause, and c4 is an emotion clause that is its own corresponding cause clause itself.
	
	In recent years, there has been a trend to use end-to-end methods to solve the ECPE task and the two sub-tasks, emotion extraction and cause extraction, which aim to extract all emotions and causes in one document. These end-to-end methods can be classified into two categories: pair-level methods by combining all clause pairs to form a clause pairing matrix; sequence labeling methods by designing novel tagging schemes. Specifically, some works ~\cite{ding2020ecpe, ding2020end, wei2020effective,chen2020end, wu2020multi} proposed pair-level methods that generate a pairing matrix by enumerating all possible combinations of clauses and then select valid emotion-cause pairs, as shown in Figure \ref{fig1:sub2}. However, the average ratio of invalid/valid pairs is more than 200:1 in the ECPE benchmark corpus \cite{xia2019emotion}, which leads to a label sparsity problem. Besides, some works \cite{chen2020unified,yuan2020emotion} proposed sequence labeling methods based on novel tagging schemes. However, they cannot effectively model the corresponding relation between emotions and causes. For example, \citet{chen2020unified} cannot deal with the situation where two emotion clauses of the same emotion type correspond to different cause clauses. Therefore, how to solve the label sparsity problem by avoiding the calculation of the pairing matrix while modeling the corresponding relation between emotions and causes is a crucial challenge. Furthermore, \citet{zhong2021frustratingly} has shown that explicit semantic information can improve performance in the extraction tasks. Therefore, utilizing explicit semantic information of clauses to improve the performance in the ECPE task is also a challenge. 
	
	To address these challenges, in this study, the ECPE task is transformed into a document-level machine reading comprehension (MRC) task, and a Multi-turn MRC framework with Rethink mechanism (MM-R) is proposed. The multi-turn structure decomposes the ECPE task to model the corresponding relation between emotions and causes and avoid generating the pairing matrix. In every turn, static and dynamic queries designed manually make full use of explicit semantic information of clauses to improve the performance of emotion or cause extraction. In addition, inspired by the human two-stage reading behavior, in which search for possible answer candidates and then verify these candidates \cite{zheng2019human}, the rethink mechanism is proposed to verify candidate emotion-cause pairs further to enhance the flow of information between emotions and causes and improve the overall performance.

	Specifically, in the first turn, all emotion clauses are extracted. In the second turn, cause clauses corresponding to each emotion clause are extracted with the assistance of explicit semantic information about emotions. Numerous experiments \cite{ding2020end,fan2020transition,ding2020ecpe,wei2020effective,chen2020end,chen2020unified,yuan2020emotion} have shown that emotion extraction is more reliable than cause extraction, and therefore it is reasonable to extract emotions first and then extract the corresponding causes with the assistance of emotions. Hence, a candidate emotion-cause pair set is obtained in the first two turns. In the third turn, a rethink mechanism verifies each candidate emotion-cause pair. For example, as shown in Figure \ref{fig1:sub3}, emotion clauses c2 and c4 can be obtained in the first turn, and cause clauses corresponding to each emotion clause can be obtained in the second turn. The candidate emotion-cause pair set \{(c2, c3), (c2, c4), (c4, c4)\} can thus be obtained without the pairing matrix. In the third turn, the rethink mechanism demonstrates that c2 is not one of the corresponding emotion clauses of c4. Therefore, the valid emotion-cause pair set becomes \{(c2, c3), (c4, c4)\}.
	
	Comprehensive experiments are conducted on the ECPE benchmark datasets to verify the effectiveness of the proposed MM-R framework. The experimental results show that the proposed framework substantially outperforms the previous methods. The contributions of this research can be summarized as follows:
	
	\begin{itemize}
		\item The ECPE task is formalized as a document-level machine reading comprehension (MRC) task. To our best knowledge, this is the first time that the ECPE task has been transferred to the MRC task.
		
		\item Based on the MRC formalization, a multi-turn MRC framework with rethink mechanism (MM-R) is proposed. It models the corresponding relation between emotions and causes and alleviates the label sparsity problem. Furthermore, the MM-R can use explicit semantic information of clauses effectively.
		
		\item The experimental results demonstrate that the proposed framework outperforms existing state-of-the-art methods.
	\end{itemize}

\section{Related Work}

	\subsection{Emotion-Cause Pair Extraction}
	
	\citet{xia2019emotion} proposed the emotion-cause pair extraction (ECPE) task and released the Chinese benchmark dataset. Recently, many methods based on the pair-level end-to-end framework have been designed for the ECPE task. For example, \citet{wei2020effective} obtained excellent results by modeling inter-clause relationships and using the ranking mechanism. \citet{ding2020ecpe} used a 2D-transformers module to model the interactions of different emotion-cause pairs. However, these pair-level end-to-end methods had a label sparsity problem caused by calculating the pairing matrix. \citet{ding2020end} used a sliding window to restrict the amount of candidate emotion-cause pairs, which aims to shrink the pairing matrix. However, the method still belonged to the pair-level end-to-end method because it assumed all clauses were emotion or cause clauses. 
	
	Sequence labeling was another popular method. However, these methods could not model the corresponding relation between emotions and causes. A unified tagging scheme \cite{chen2020unified} could not be used on a document with multiple emotion clauses of the same emotion types because the tagging scheme was based on emotion types. Another joint tagging scheme \cite{yuan2020emotion} based on the distance between the cause and the corresponding triggered emotion. The tagging scheme could not handle the situation in which the distance between the emotion clause and the corresponding cause clause exceeded the distance threshold. Therefore, the challenge remains of how to solve the label sparsity problem while modeling the corresponding relation between emotions and causes. 
	
	\subsection{Machine Reading Comprehension}
	
	Recently, it became popular to transform many traditional natural language processing tasks into machine reading comprehension (MRC) tasks. %\cite{li2019entity} used a multi-turn question answering (QA) approach to accomplish entity-relation extraction.
	For example, \citet{liu2020event} cast the event extraction as a machine reading comprehension problem to solve the data scarcity problem. \citet{zhou2021role} proposed a dual QA framework aiming at event argument extraction. \citet{chen2021bidirectional} and \citet{mao2021joint} proposed a bidirectional MRC and a Dual-MRC frameworks to handle aspect-based sentiment analysis, respectively. In this paper, the ECPE task is formalized as a document-level MRC task, and the multi-turn MRC framework with rethink mechanism is proposed. It concatenates a query with each clause to explicitly utilize semantic information of clauses and uses a single classifier to predict whether a clause is an emotion or a cause clause in each turn.

	%In this paper, we formalize the ECPE task as a document-level MRC task and propose the multi-turn MRC framework with rethink mechanism. Different from the aforementioned works which essentially belong to token-level binary classification tasks, the ECPE task is a document-level clause pair classification task and we transform the ECPE task to a 

\section{Methodology}
	
	\subsection{Problem Formulation}
	Given a document consisting of multiple clauses $D = (c_1, c_2, ..., c_{|D|})$ \footnote{The $|*|$ denotes the number of elements in the collection $*$.} and each clause contains multiple words $c_i=(w_{i,1},w_{i,2}, ...,w_{i,|c_i|})$, our goal is to extract a set of emotion-cause pair in $D$:
	\begin{equation}
		P= \{(c_k^{e_i}, c^{ca_{i,j}}_k)\}_{k=1}^{|P|},
	\end{equation}
	where the superscript $e_i$ and $ca_{i,j}$ denote the $i$-th emotion clause and its corresponding $j$-th cause clause in $D$; the subscript $k$ denotes the $k$-th emotion-cause pair in set $P$; an emotion clause corresponds to one or more cause clauses. \footnote{To avoid confusion, the ``cause clause'' is denoted to $ca$ rather than $c$.}
	%where $c^{e_i}$ is the $i$-th emotion clause in $D$, $c^{c_{i,j}}$ :is the corresponding $j$-th cause clause of emotion clause $c^{e_i}$, and $(c_k^{e_i}, c_k^{c_{i,j}})$ is the $k$-th emotion-cause pair in set $P$.
	%\footnote{An emotion clause corresponds to one or more cause clauses, and vice versa.}	
	
	\subsection{Query Design}
	Static and dynamic queries are used to formalize the ECPE task to the MRC task. All queries can be formulated as follows:\footnote{Static cause query and static pair query are only used in two variants of our framework (MM-D and MRC-E2E); refer to the Experiments section for details.}

	\begin{itemize}
		\item \textbf{Static emotion query} $q^{se} \in Q^{se}$: The query ``\textit{Is it an emotion clause?}'' is designed to extract all emotion clauses.
		
		\item \textbf{Static cause query} $q^{sc} \in Q^{sc}$: The query ``\textit{Is it a cause clause?}'' is designed to extract all cause clauses.
		
		\item \textbf{Static pair query} $q^{sp} \in Q^{sp}$: The query ``\textit{Is it an emotion-cause pair?}'' is designed to extract all emotion-cause pairs.
		
		\item \textbf{Dynamic emotion query} $q^{de} \in Q^{de}$: The query template ``\textit{Is it an emotion clause corresponding to $c_i$?}'' is designed to extract emotion clauses corresponding to clause $c_i$.
		
		\item \textbf{Dynamic cause query} $q^{dc} \in Q^{dc}$: The query template ``\textit{Is it a cause clause corresponding to $c_i$?}'' is designed to extract cause clauses corresponding to clause $c_i$.
	\end{itemize}
	
	%The first two turns generate candidate emotion-cause pairs, and the third turn performs backward verification. And for the Chinese corpus, these questions are constructed in Chinese.
	
	\begin{figure*}[htb]
    	\centering
    	\includegraphics[width=1.0\linewidth]{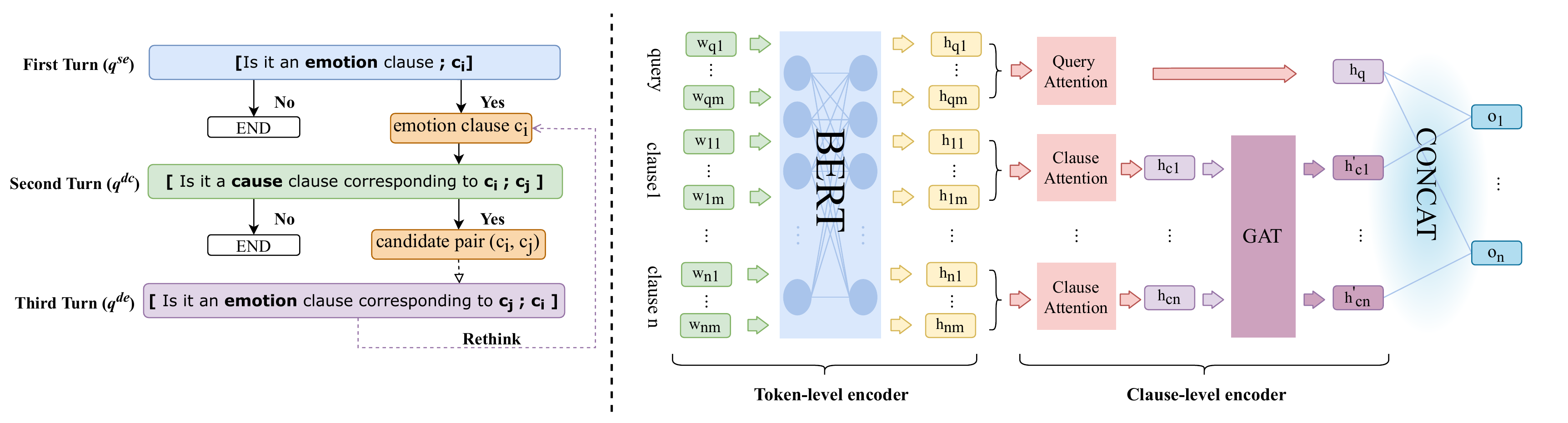}
    	\caption{\textit{Left}: The overall architecture of our MM-R framework. In each turn, the answer is yes if the probability output by the classifier is greater than 0.5, otherwise it is no. \textit{Right}: The implementation structure of the encoding layer which includes the token-level encoder and the clause-level encoder. The token-level encoder generates the hidden representation of each token using the BERT module. The clause-level encoder provides the hidden representation of query and each clause using the attention mechanism and graph attention network. Finally, the concatenate operation (CONCAT) is executed on the hidden representations of queries and clauses.}
    	\label{figure:main}
    \end{figure*}
	
	% The MM-R framework includes generating the set of candidate emotion-cause pairs in the first two turns and verifying candidate pairs in the third turn. Specifically,

	\subsection{Frameworks}
	The architecture of MM-R is illustrated on the left side of Figure \ref{figure:main}. In the first turn, a static emotion query $q^{se}$ is used to extract emotion clause set $E=\{c^{e_i}\}_{i=1}^{|E|}$. In the second turn, based on each extracted emotion clause $c^{e_i} \in E$, a dynamic cause query $q^{dc}$ is constructed to obtain $c^{e_i}$'s corresponding cause clause set $C_i = \{c^{ca_{i,j}}\}_{j=1}^{|C_{i}|}$. In this way, the candidate emotion-cause pair set $P^{can} = \{(c_k^{e_i}, c_k^{ca_{i,j}})\}_{k=1}^{|P^{can}|}$ is obtained in the first two turns. To filter out invalid emotion-cause pairs in $P^{can}$ further, for each cause clause $c^{ca_{i,j}}$, the framework rethinks whether the clause $c^{e_i}$ is its corresponding emotion clause with the help of dynamic emotion query $q^{de}$. Finally, the valid emotion-cause pair set $P = \{(c_k^{e_i}, c_k^{ca_{i,j}})\}_{k=1}^{|P|}$ is obtained. 
	
	\subsection{Encoding Layer}
	The structure of the encoding layer is illustrated on the right side of Figure \ref{figure:main}. The encoding layer includes the token-level encoding layer and the clause-level encoding layer. 
	
	\textbf{Token-level encoder}: This study used BERT \cite{devlin2019bert} as its token-level contextualized encoder. Specifically, a document $D = (c_1, c_2, ..., c_{|D|})$ and a query $q$ are used to construct the BERT input sequence:
	\begin{equation}
		\begin{split}
			I=\{[CLS],w_{q,1}, w_{q,2}, ..., w_{q,|q|},[SEP],\\
			w_{1,1}, w_{1,2}, ..., w_{|D|,1}, ..., w_{|D|, |c_{|D|}|}\},
		\end{split}
	\end{equation}
	where $q=q^{se}$ in the first turn, $q=q^{dc}$ in the second turn and $q=q^{de}$ in the third turn; $w_{q,j}$ is the $j$-th token of query $q$; $w_{i,j}$ is the $j$-th token of the $i$-th clause in the document $D$; and $[CLS], [SEP]$ are special BERT tokens. The tokens can then be encoded into hidden representations:
	\begin{equation}
		\begin{split}
			H^I= &BERT(I) \\
			=&\{h_{[CLS]},h_{q,1}, h_{q,2}, ..., h_{q,|q|},h_{[SEP]},\\
			&\ \  h_{1,1},  h_{1,2},..., h_{|D|,1}, ..., h_{|D|, |c_{|D|}|}\},
		\end{split}
	\end{equation}
	where $H^I \in \mathbb{R}^{|I| \times d}$; $d$ is the dimension of the hidden states; and $h_{i,j}$ denotes the hidden representation of token $w_{i,j}$. 
	
	\textbf{Clause-level encoder}: The attention mechanism and the graph attention network (GAT) \cite{velivckovic2017graph} are used to obtain clause-level hidden representations. 
	
	Specifically, for clause $c_i$, its token-level representation set is selected from $H^I$:
	\begin{equation}
		S_{c_i} = \{h_{i,j}\}_{j=1}^{|c_i|} \in \mathbb{R}^{|c_i| \times d}.  
	\end{equation}
	
	The attention mechanism produces an attention weight vector $\alpha_i$. The hidden representation of clause $c_i$ is obtained by calculating the weighted sum of the hidden representations of all tokens:
	\begin{equation}
		\alpha_i = softmax(w^TS_{c_i} + b)  \in \mathbb{R}^{1 \times |c_i|},
	\end{equation}
	\begin{equation}
		h_{c_i}  = sum(\alpha_i S_{c_i}) \in \mathbb{R}^{1 \times d},
	\end{equation}
	where $w$ and $b$ are learnable parameters. For all clauses, the attention mechanism is used to obtain the hidden representations:
	\begin{equation}
		H_C = \{h_{c_1}, h_{c_2},..., h_{c_{|D|}}\}.    
	\end{equation}
	
	Similarly, the attention mechanism is used to obtain the hidden representation of query $q$:
	\begin{equation}
		H_Q = \{h_q\}.
	\end{equation}
	
	Furthermore, GAT models the interaction among clauses, and then the hidden representations of clauses are updated to:
	\begin{equation}
	    H'_{C} = GAT(H_{C})
		 = \{h'_{c_1}, h'_{c_2}, ..., h'_{c_{|D|}} \}.
	\end{equation}
	
	Finally, $h_{q} \in H_Q$ and $h'_{c_i} \in H_C^{'}$ are concatenated to obtain $o_{i} = [h_{q}; h'_{c_i}]$, where $[ ; ]$ denotes the concatenate operation. The output of the encoding layer is thus obtained:
	\begin{equation}
		O_{enc} = \{o_1, o_2, ..., o_{|D|} \}.
	\end{equation}
	
	\subsection{Answering Prediction}
	
	The output representation of the encoder layer $o_{i} = [h_{q}; h'_{c_i}]$ contains abundant semantic information, including identification information of emotion/cause, semantic information of clauses, and semantic information of the document. Therefore, only a single linear perception (SLP) is needed to obtain the answer to the query by predicting yes or no:
	\begin{equation}
		\begin{split}
			\hat{y}_i = \sigma(w_S^To_i + b_S),
		\end{split}
	\end{equation}
	where $w_S$ and $b_S$ are learnable parameters of SLP; $\sigma(\cdot)$ is a logistic function; and $\hat{y}_i$ denotes the probability that the answer is yes. If $\hat{y}_i > 0.5$, the answer is judged to be yes, meaning that clause $c_i$ is one of the answers to query $q$. This prediction style is similar to BoolQ \cite{clark2019boolq}.

	\subsection{Joint Training}
	A joint learning strategy was used in this study, the cross-entropy loss for each type of query was minimized as follows:
	\begin{equation}
		\begin{split}
			& \mathcal{L}^{*} = \\ & -\sum_{i=1}^{N}\sum_{j=1}^{|D|}\sum_{k=1}^{|Q^{*}|} [p(y_{i,j,k} | c_{i,j},q_k^{*}) \log \hat{p}(y_{i,j,k} | c_{i,j},q_k^{*})],
		\end{split}
	\end{equation}
	where $* \in \{se, dc, de\}$; $ N$ denotes the number of documents in the dataset; $c_{i,j}$ is the $j$-th clause of the $i$-th document; and $q_k^{*}$ is the $k$-th query in $Q^{*}$. 
	
	Therefore, the final loss function $\mathcal{L}$ was as follows:
	\begin{equation}
		\mathcal{L} = \mathcal{L}^{se} +  \mathcal{L}^{dc} + \mathcal{L}^{de},
	\end{equation}

	\subsection{Inference}
	
	In the first turn, the $q^{se} \in Q^{se}$ first identifies the emotion clause set $E = \{c^{e_1}, c^{e_2}, ..., c^{e_{|E|}}\}$. In the second turn, for each predicted emotion clause $c^{e_i}$, the $q^{dc} \in Q^{dc}$ recognizes the corresponding cause clause set $C_i = \{c^{ca_{i,1}},c^{ca_{i,2}}, ..., c^{ca_{i,|C_i|}}\}$ and obtains the set of candidate emotion-cause pairs $P^{can} = \{(c_k^{e_i}, c_k^{ca_{i,j}})\}_{k=1}^{|P^{can}|}$. And the probability of each candidate pair is $p(c^{e_i}, c^{ca_{i,j}}) = p(c^{e_i})p(c^{ca_{i,j}}|c^{e_i})$. 
	
	In the third turn, the $q^{de} \in Q^{de}$ is used for selecting valid emotion-cause pairs. Furthermore, the rethink mechanism is implemented through a soft selection strategy, in which the probability is adjusted and a probability threshold is used. Specifically, the probability of the candidate pair $(c^{e_i}, c^{ca_{i,j}})$ is updated by this strategy as follows:
		\begin{equation}
			p(c^{e_i}, c^{ca_{i,j}}) = \lambda p(c^{e_i})p(c^{ca_{i,j}}|c^{e_i}), 
		\end{equation}
	where the weight factor $\lambda$ is used to adjust the probability of candidate emotion-cause pairs. $\lambda$ is 1 when the predicted result of the third turn is yes, otherwise $\lambda$ is a unique value between 0 and 1. The set of valid emotion-cause pairs is as follows:
		\begin{equation}
		\begin{split}
			& P = \\
			& \{ (c^{e_i}, c^{ca_{i,j}}) | (c^{e_i}, c^{ca_{i,j}}) \in P^{can}, p(c^{e_i}, c^{ca_{i,j}}) > \delta \},
			\end{split}
		\end{equation}
	where $\delta$ is a probability threshold value.

	% Table generated by Excel2LaTeX from sheet 'Sheet2'
	\begin{table*}
		\centering
		\begin{tabular}{ccccccccccccc}
			
			\toprule[1.5pt]
			\multirow{2}{*}{{\textbf{Model}}} & & \multicolumn{3}{c}{E-C Pair Extraction} & & \multicolumn{3}{c}{Emotion Extraction}&  & \multicolumn{3}{c}{Cause Extraction} \\
			\cmidrule{3-5} \cmidrule{7-9} \cmidrule{11-13}& & P(\%)  & R(\%)   & F1(\%)  && P(\%)   & R(\%)  & F1(\%)  &  & P(\%)     & R(\%)  & F1(\%) \\
			\midrule
			%L-J-MANN & & 71.10  & 60.70  & 65.50  & & 89.90  & 80.00    & 84.70  & & -     & -     & - \\
			SL-NTS & & 72.43 & 63.66 & 67.76  & & 81.96 & 73.29 & 77.39  & & 74.90  & 66.02 & 70.18 \\	
			TransDGC (Val) & & 73.74 & 63.07 & 67.99  & & 87.16 & 82.44 & 84.74  & & 75.62  & 64.71 & 69.74 \\	
			ECPE-2D&  & 72.92 & 65.44 & 68.89 & & 86.27 & \textbf{92.21} & 89.10  & & 73.36 & 69.34 & 71.23 \\
			PairGCN & & 76.92 & 67.91 & 72.02 & & 88.57 & 79.58 & 83.75 & & 79.07 & 69.28 & 73.75 \\
			RANKCP & & 71.19 & 76.30  & 73.60  & & 91.23 & 89.99 & 90.57 & & 74.61 & 77.88 & 76.15 \\
			ECPE-MLL & & 77.00    & 72.35 & 74.52 & & 86.08 & 91.91 & 88.86 & & 73.82 & 79.12 & 76.30 \\
			
			%IE & & 71.49 & 62.79 & 66.86 & &  86.14 & 78.11 & 81.88  &&  73.48  & 58.41  & 64.96 \\
			\midrule
			MM-R  & & \textbf{82.18} & \textbf{79.27} & \textbf{80.62} & & \textbf{97.38} & 90.38  & \textbf{93.70}  && \textbf{83.28} & \textbf{79.64} & \textbf{81.35} \\	
			MM-R (Val)  & & 78.97 & 75.32 & 77.06 & & 96.09 & 88.09  & 91.88  && 80.90 & 76.21 & 78.45 \\	
			\bottomrule[1.5pt]		
		\end{tabular}%
		\caption{Performance of our models and baselines. P, R and F1 denote precision, recall and F1-measure respectively. E-C denotes Emotion-Cause. TransDGC(Val) and MM-R(Val) use the second data split style, the rest of models use the first data split style.}
		\label{tab:exp}%
	\end{table*}
	
		%MM-R uses the same data split with the previous works. %MM-R-Dev uses the data split of dividing the corpus into a training/development/test set in a ratio of 8:1:1, so its results are more reliable. However, to be fair, the results of MM-R are used in the following analysis of experimental results.

\section{Experiments}
	
	\subsection{Dataset and Metrics}
	
	The benchmark dataset \cite{xia2019emotion} was constructed based on a public Chinese emotion corpus \cite{gui2016event} from the SINA NEWS website \footnote{https://news.sina.com.cn/}. The dataset contains 1,945 documents and 28,727 clauses. The number of candidate clause pairs is 490,367 and the number of valid emotion-cause clause pairs is 2,167.
 
 %The dataset contains 1,945 documents and 28,727 clauses. The number of candidate clause pairs is 490,367 and the number of valid emotion-cause clause pairs is 2,167.
	
	In the experiment, we use the two styles of data split:
 
    \begin{itemize}
		\item \textbf{10-fold cross validation}. Selecting 90\% of the data for training and the remaining 10\% for testing stochastically (as same as \citet{xia2019emotion}). The most previous works use the data split style.

        \item \textbf{Training/Validation/Test data set}. Selecting 80\% of the data for training, 10\% of the data for validating and the remaining 10\% for testing stochastically (as same as \citet{fan2020transition}). The data split style is more plausible than the first style.
  \end{itemize}
 
 %Moreover, to make the results more reliable, another data split as same as \cite{fan2020transition} was used (stochastically divided the corpus into a training/development/test set in a ratio of 8:1:1). We repeated the experiments 20 times and reported the average results.
	
	Furthermore, when we extract the emotion-cause pairs, we obtain the emotions and causes simultaneously. Thus, the performance of emotion extraction and cause extraction were also evaluated. The precision P, recall R and F1 score defined in \cite{gui2016event,xia2019emotion} are used to evaluate the performance of the three tasks.

	\subsection{Experimental Settings}
	
	We used the BERT$_{base-Chinese}$ as our encoding backbone. During training, the AdamW optimizer with a weight decay of 0.01 was used for online learning, and the initial learning rate and warmup rate are set to 1e-5 and 0.1 respectively. The batch size is set to 2. As for regularization, dropout is applied for networks and the dropout rate is set to 0.1. We trained the model 20 epochs in total and adopted early stopping strategy. In the inference stage, $\lambda \in \{0.7, 1.0\}$ and the threshold $\delta$ was set to 0.5. The model was run on a Tesla V100 GPU.

	\subsection{Baselines}
	To demonstrate the effectiveness of our method, we compare our model with the following BERT baselines.
	
	\begin{itemize}
		%\item \textbf{LAE-JOINT-MANN} %\cite{tang2020joint} explores a multi-level attentional module and captures the implicit connection between the emotion detection task and  the ECPE task.
		\item \textbf{SL-NTS} \cite{yuan2020emotion} regards the ECPE task as a sequence labeling problem and proposes a novel tagging scheme.
		\item \textbf{TransDGC} \cite{fan2020transition} proposes a transition-based model to transform the task into a parsing-like directed graph construction procedure. \footnote{It should be noted that the TransDGC method use the second data split style, while others methods use the first data split style.}
		\item \textbf{ECPE-2D} \cite{ding2020ecpe} uses a 2D transformers to model the interactions of different emotion-cause pairs. 
		\item \textbf{PairGCN} \cite{chen2020end} constructs a pair graph convolutional network to model dependency relations among local neighboring candidate pairs.
		\item \textbf{RANKCP} \cite{wei2020effective} uses a graph attention network to model interactions between clauses and selects emotion-cause pairs by the ranking mechanism. 
		\item \textbf{ECPE-MLL} \cite{ding2020end} extracts emotion-cause pairs based on sliding window multi-label learning. It is a state-of-the-art model of baselines. 
		
		%\item \textbf{IE} \cite{chen2020unified} assigns emotion type labels to emotion and cause clauses and proposes a unified sequence labeling model.
		
	\end{itemize}

	\begin{table*}
		\centering
		\begin{tabular}{cccc}
			\toprule[1.5pt]
			 & Natural QL & Pseudo QL &  Structured QL \\
			\midrule
			$Q^{se}$   & \textit{Is it an emotion clause?} & \textit{emotion?} & \textit{emotion:\_;cause:None} \\	
			$Q^{dc}$   & \textit{Is it a cause clause corresponding to $c_i$?} & \textit{$c_i$;cause?} & \textit{emotion:$c_i$;cause:\_} \\
			$Q^{de}$   & \textit{Is it an emotion clause corresponding to $c_i$?} & \textit{$c_i$;emotion?} & \textit{emotion:\_;cause:$c_i$} \\
			\midrule
			MM-R   & 80.62 (\%F1)  & 80.51 (\%F1) & 79.72 (\%F1) \\
			\bottomrule[1.5pt]
			
		\end{tabular}%
		\caption{The performance of different query language designs (Natural, Pseudo and Structured QL) on ECPE task. ``QL" denotes ``Query Language". $Q^{se}, Q^{dc}$ and $Q^{de}$ are static emotion query, dynamic cause query and dynamic emotion query, respectively.}
		\label{table:queryDesign}%
	\end{table*}

	\subsection{Variants}
	To show the effectiveness of the multi-turn structure and the rethink mechanism in the proposed method, We designed the following variants:
	\begin{itemize}
		\item \textbf{MRC-E2E} is a single-turn \textbf{MRC} framework with an \textbf{E}nd-\textbf{t}o-\textbf{E}nd style. The static queries $q^{sp}$ is used to extract an emotion-cause pair set.
		\item \textbf{MM} is a simple \textbf{M}ulti-turn \textbf{M}RC framework. Compared with MM-R, it removes the rethink mechanism. 
		\item \textbf{MM-D} is proposed inspired by \cite{chen2021bidirectional, mao2021joint}, which is a \textbf{M}ulti-turn \textbf{M}RC framework with \textbf{D}ual structure. One direction sequentially recognizes emotions and causes to obtain the candidate emotion-cause pair set with the help of static emotion query $q^{se}$ and dynamic cause query $q^{dc}$. In contrast, the other direction identifies causes first and then emotions to obtain another candidate set with the help of static cause query $q^{sc}$ and dynamic emotion query $q^{de}$. Finally, we take the intersection of the two candidate sets.
	\end{itemize}

	\subsection{Main Results}

	Table \ref{tab:exp} gives the comparative results for the ECPE task and the two sub-tasks. The proposed MM-R shows a clear advantage over other baselines, obtaining $F_1$ improvements of 6.10\%, 3.13\%, and 5.05\%, respectively, over the previous baselines on the three tasks. 

	Specifically, for the ECPE task, the MM-R obtains 5.18\%, 6.92\% and 6.10\% improvements in the $P$, $R$, and $F1$ measures compared to ECPE-MLL. For the emotion extraction task, the MM-R performs 3.13\% better than the best-performed baseline RANKCP, and it is worth noting that the increase in precision contributed most to the boost in the $F_1$ score. %It can conclude that the rethink mechanism in the third turn made the precision better because some negative samples were filtered out.
	We believe that the high precision is due to the rethink mechanism filtering out some negative samples. The cause extraction task is more difficult because the cause clauses depend heavily on emotions. However, with the help of dynamic cause queries constructed by emotion semantic information, substantial increases (+5.05\% $F_1$) are obtained compared to ECPE-MLL. 
	
	Besides, when we use the data split as the TransDGC \cite{fan2020transition}, our method MM-R (Val) still obtains the state-of-the-art performance over the previous baselines.

	\begin{figure*}[htb]
		\centering
		\includegraphics[width=2.0\columnwidth]{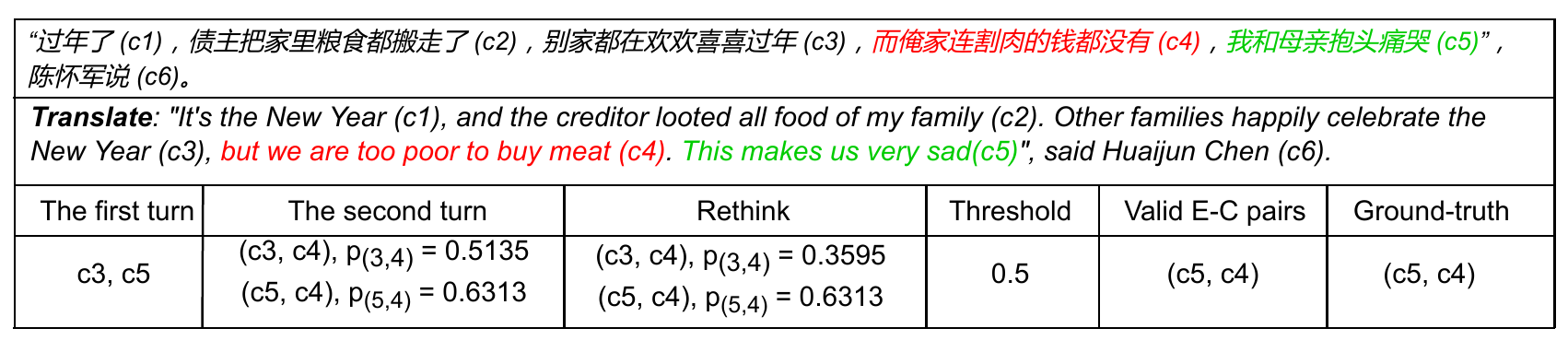}
		\caption{An example in the test set. The emotion clause set \{c3, c5\} was obtained in the first turn and the candidate emotion-cause pair set \{(c3, c4), (c5, c4)\} in the second turn. After using the rethink mechanism, the valid emotion-cause pair set was identified as \{(c5, c4)\}.}
		\label{figure:case}
	\end{figure*}

\section{Further Analysis}
	
	\subsection{Effect of the Multi-turn Structure}
	
	MRC-E2E has the same label sparsity problem as pair-level end-to-end baselines, and it cannot model the interaction of emotions and causes due to its single-turn structure.  %compared with other three multi-turn MRC frameworks, it obtains only 90.34\%, 77.92\%, and 75.35\% $F_1$ scores. 
	However, MM, the most straightforward multi-turn framework among these similar methods, achieves 2.68\%, 0.02\%, and 2.84\% $F_1$ improvements over MRC-E2E as shown in Table \ref{table:variants}, which demonstrates the effectiveness of the multi-turn structure. 
	
	In addition, compared with end-to-end baseline ECPE-MLL, MRC-E2E still achieves 1.48\%, 1.62\% and 0.83\% improvements of $F_1$ in three tasks, which demonstrates that the explicit emotion/cause semantic information played an important role. 
	
	\begin{table}[tbp]
		\centering
		\begin{tabular}{cccc}
			\toprule[1.5pt]
			\multirow{2}[0]{*}{\textbf{Model}} & \multicolumn{3}{c}{Extraction of. (F1 \%)} \\
			\cmidrule{2-4}
			& Emotion & Cause & E-C pair \\
			\midrule
			MRC-E2E & 90.34 & 77.92 & 75.35 \\
			MM & 93.02 & 77.94 & 78.19 \\
			MM-D & 93.67 & 79.47 & 78.76 \\
			MM-R & \textbf{93.70} & \textbf{81.35} & \textbf{80.62} \\
			\bottomrule[1.5pt]
		\end{tabular}
		\caption{Performance of variants. }
		\label{table:variants}
	\end{table}

	\subsection{Effect of the Rethink Mechanism}
	To verify the advantages of the rethink mechanism, MM-R was compared with the other two models (MM and MM-D). Specifically, compared with MM, MM-R achieved 0.68\%, 3.41\%, and 2.43\% F1 improvements on the three tasks, demonstrating the the rethink mechanism's effectiveness. Furthermore, MM-D achieved excellent performance with its dual structure. However, its score was 1.86\% lower than MM-R in the ECPE task. The dual structure treats both emotion extraction and causes extraction equally, whereas the rethink mechanism is more trusting of emotion extraction. MM-R obtains a better performance due to emotion extraction being more reliable than cause extraction\footnote{We have illustrated in the introduction that emotion extraction is more reliable than cause extraction.}.

\subsection{Effect of the Query Design}

We explore the effect of different query language design on the model. As shown in Table \ref{table:queryDesign}, their F1 scores all exceeded 79\% on the ECPE task. Concretely, Natural QL performs best in the ECPE task. Compared with Pseudo and Structured QL, Natural QL makes it easier for our model to understand the meaning of the queries, because BERT's pre-trained corpus are all natural languages. The Pseudo QL can be regarded as a concise expression of Natural QL. Hence our model can still understand the meaning of queries. Structured QL is the most abstract of three QLs, transforming the three queries ($Q^{se, dc, de}$) into a unified paradigm. It gains the lowest F1 score, which indicates that the inductive ability of our model is not enough to understand abstract structured query language.

	\subsection{Case Analysis}
	Figure \ref{figure:case} shows the inference process of MM-R and the advantage of the rethink mechanism through an example in the test set. The candidate emotion-cause pair set \{(c3, c4), (c5, c4)\} were obtained in the first two turns, and their probabilities were $p_{3, 4} = 0.5135$ and $p_{5, 4}=0.6313$. After this, the proposed model adjusted the probabilities by rethinking whether each candidate emotion-cause pair was valid. The bottom of Figure \ref{figure:case} shows that the adjusted probabilities were $p_{3, 4} = 0.3595$ and $p_{5, 4} = 0.6313$, and the threshold value was 0.5. This means that the answer to the dynamic emotion query \textit{``Is it an emotion clause corresponding to `but we are too poor to buy meat'?"} was no for clause $c3$ in the third turn. Therefore, candidate emotion-cause pair (c3, c4) was filtered out, while (c5, c4) was preserved.

	\subsection{Results for Emotion Cause Extraction}
	The proposed framework is also applied to the emotion cause extraction (ECE) task. Compared with the ECPE task, the ECE task has emotion annotations, and therefore in the MM-R framework, the emotion extraction of the first turn is omitted, and the emotion annotations are used to extract the corresponding cause clauses directly in the second turn. Finally, each candidate emotion-cause pair was verified by the rethink mechanism. 
	
    Comparisons are also conducted with recently proposed methods for the ECE task: RTHN \cite{xia2019rthn}, RHNN \cite{fan2019knowledge}, KAG \cite{yan2021position} and 2-step RANKING \cite{xu2021two}. Table \ref{tab:ece} clearly demonstrates that our proposed framework achieves the state-of-the-art performance on the ECE task.

    \begin{table}[tbp]
          \centering
            \begin{tabular}{cccc}
            \toprule[1.5pt]
            \multirow{2}[4]{*}{\textbf{Methods}} & \multicolumn{3}{c}{Emotion Cause Extraction} \\
        \cmidrule{2-4} & P(\%) & R(\%) & F1(\%) \\
            \midrule
            %PAE-DGL & 76.19 & 69.08 & 72.42 \\
            RTHN  & 76.97 & 76.62 & 76.77 \\
            KAG & 79.12 & 75.81 & 77.43 \\
            RHNN  &81.12 & 77.25 & 79.14 \\
            2-step RANKING & 80.76 & 78.45 & 79.59 \\
            \midrule
            MM-R  & \textbf{83.59} & \textbf{83.47} & \textbf{83.48} \\
            \bottomrule[1.5pt]
            \end{tabular}
            
          \caption{Results on the Emotion Cause Extraction task.}
          \label{tab:ece}
    \end{table}%

\section{Conclusions}
	
	This study transforms the emotion-cause pair extraction (ECPE) task into the machine reading comprehension (MRC) task and proposes a multi-turn MRC with rethink mechanism (MM-R). %Specifically, the static emotion query and the dynamic cause query are designed to extract emotion clauses and the corresponding cause clauses.
	This structure, which extracts emotions and causes in turn, avoids the label sparsity problem and models the complicated corresponding relations between emotions and causes. In every turn, explicit semantic information can be used effectively. Furthermore, the rethink mechanism verifies each candidate emotion-cause pair by modeling the flow of information from causes to emotions. Experimental results on the ECPE corpus demonstrated the effectiveness of the proposed model.
	
% 	In future work, we will explore the performance of our method in multi-lingual and cross-lingual scenarios. Besides, the emotion type should be taken into account in the ECPE task. Fine-grained emotions contribute to extracting the cause clauses, especially in a document with multiple emotions.

\section*{Acknowledgements}
This work was supported by National Key Research and Development Program of China (Grant No. 2020YFC0833402), National Natural Science Foundation of China (Grant Nos. 61976021, U1811262), and Beijing Academy of Artificial Intelligence (BAAI).

\bibliographystyle{acl_natbib}
\bibliography{myref}

\end{document}